\documentclass[10pt,twocolumn,letterpaper]{article}

\usepackage{wacv}
\usepackage{times}
\usepackage{epsfig}
\usepackage{graphicx}
\usepackage{amsmath}
\usepackage{amssymb}

\usepackage{comment}
\usepackage{amsmath,amssymb} 
\usepackage{color}
\usepackage{soul}
\usepackage{makecell}
\usepackage{booktabs}
\usepackage[export]{adjustbox}

\usepackage[hoptionsi]{overpic}
\usepackage{array}
\usepackage{caption}
\usepackage{subcaption}
\usepackage{float}
\usepackage{pdfpages}

\wacvfinalcopy 

\ifwacvfinal
\fi

\ifwacvfinal
\usepackage[breaklinks=true,bookmarks=false]{hyperref}
\else
\usepackage[pagebackref=true,breaklinks=true,colorlinks,bookmarks=false]{hyperref}
\fi

\usepackage[utf8]{inputenc}

\begin{document}

\title{Neural Alignment for Face De-pixelization}

\author{Ma'ayan Shuvi\textsuperscript{1}, Noa Fish\textsuperscript{2}, Kfir Aberman\textsuperscript{1}, Ariel Shamir\textsuperscript{2}, Daniel Cohen-Or\textsuperscript{1} \\
\\
  \textsuperscript{1}Tel-Aviv University, Tel-Aviv, Israel\\
  \textsuperscript{2}The Interdisciplinary Center, Herzliya, Israel\\
}

\maketitle
\begin{abstract}

We present a simple method to reconstruct a high-resolution video from a face-video, where a person’s identity is obscured by pixelization. This concealment method is popular because the viewer can still perceive a human face figure and the overall head motion. However, we show in our experiments that a fairly good approximation of the original video can be reconstructed in a way that compromises anonymity. Our system exploits the simultaneous similarity and small disparity between close-by video frames depicting a human face, and employs a spatial transformation component that learns the alignment between the pixelated frames. 
Each frame, supported by its aligned surrounding frames, is first encoded, then decoded to a higher resolution. Reconstruction and perceptual losses promote adherence to the ground-truth, and an adversarial loss assists in maintaining domain faithfulness. There is no need for explicit temporal coherency loss as it is maintained implicitly by the alignment of neighboring frames and reconstruction.  
Although simple, our framework synthesizes high-quality face reconstructions, demonstrating that given the statistical prior of a human face, multiple aligned pixelated frames contain sufficient information to reconstruct a high-quality approximation of the original signal.
\end{abstract}

\noindent

\section{Introduction}

One of the most popular methods to hide a person’s identity in a video is to pixelize his or her face. Using this method, the viewer can still perceive a human face figure and the overall head motion, and sometimes also recognize general aspects such as gender or complexion. These aspects promote the reliability of the video compared, for example, to totally covering the face region. However, as we show in this paper, using neural networks, we can reconstruct from such pixelated video, a fairly good approximation of the original video where anonymity is compromised. 
Our results demonstrate that given the statistical prior of a human face, a temporal set of frames depicting a pixelated moving head often contain enough information to fairly reconstruct the subject's identity, facial expression and head pose.

\begin{figure*}
\captionsetup{labelfont=bf}
\newcommand{\plfig}{11.5}
\centering

\begin{overpic}[width=\plfig cm]{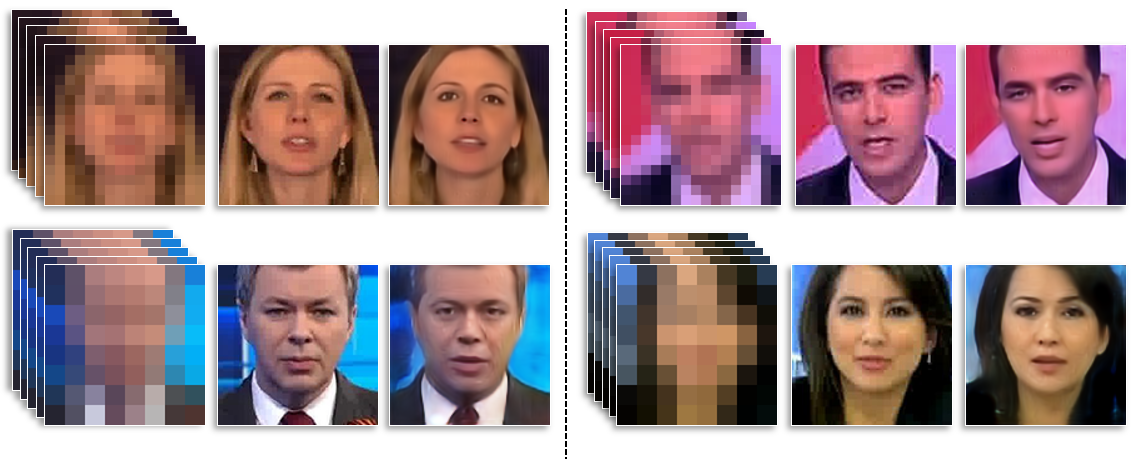}
    \put(8, 0.5){\textcolor{black}{input}}
    \put(24, 0.5){\textcolor{black}{GT}}
    \put(38, 0.5){\textcolor{black}{output}}
    \put(59, 0.5){\textcolor{black}{input}}
    \put(75, 0.5){\textcolor{black}{GT}}
    \put(88.5, 0.5){\textcolor{black}{output}}
\end{overpic}

\caption{Four sequences of pixelated frames. The results of our method exhibit a fair approximation to the groundtruth. Top row pixelization is 16x16, while bottom row is 8x8.}
\label{fig:teaser}
\end{figure*}
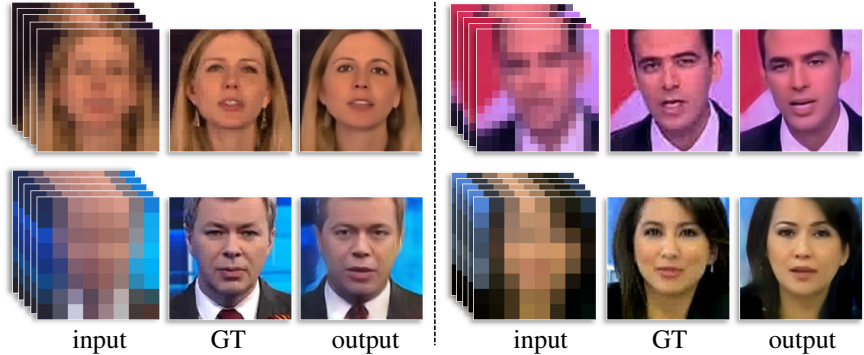

Our method relates to many previous works that deal with super resolution \cite{yang2008image,yang2010image,dong2015image}, but we deliberately choose to concentrate on the specific domain of faces. 
{Adopting the general paradigm of alignment and reconstruction, our solution falls within the realm of many successful previous techniques}
\cite{Yoshida2012RobustFS, wang2019edvr, Cansizoglu2018SuperresolutionOV}
{, but, presents a clean and simple take on this well-utilized approach, by combining the two steps along with temporal information, without compromising performance.}
By focusing our efforts on a homogeneous dataset that is not only class-specific, but also highly structured, we are able to employ simple and effective learning mechanisms unburdened by data generality characterized by a large variance.
{Our method is thus simple and intuitive, combining deep spatial alignment in place of facial landmark detection, an important distinction when one strives to handle highly low resolution images depicting faces, where landmark detection becomes intractable.}
Focusing on faces also relates to many recent works that reconstruct a 3D model followed by a rendering step and composition back to the video \cite{Face2Face2016,Tweari2019,Deng2019,Fried:2019}.
These methods are better suited, however, in situations where the input is expected to be of only a moderately low resolution, and is less applicable for substantially pixelated data. 
We utilize the well-defined and structured prior, where there is no  significant change in view angle or in the position of parts (eyes, nose, mouth).
Such data is still informative enough to power commonality detection among its instances, particularly among close-by frames extracted from a clip. These detected commonalities are a key component in our method, as they offer a potential boost of information to an otherwise minimal image of a face.

We design our solution as a combination of two main parts, and employ a many-to-one paradigm, where a single low resolution frame is supported by its surrounding frames. To better exploit the information from potentially misaligned frames, we first train a spatial transformation network (STN) \cite{jaderberg2015spatial} to learn the warp from one frame to another. The trained STN is then used per input frame, to predict a free-form warp grid from each of its supporting frames, thereby facilitating the formation of a temporally aligned stack of images. 
{The free-form field supports non-rigid deformations which are crucial for alignment of animate objects such as human faces, and acts as an alternative to the commonly used warp driven by corresponding facial landmarks, which we deliberately wish to avoid.}

The aligned frames are passed on to the second part -- an encoder-decoder that outputs the HR version of the input frame. The auto-encoder consists of several convolutional layers with skip-connections. Previous works demonstrated the ability to synthesize high-resolution frames of face and body using such a structure \cite{kim2018deep,aberman2019deep,chan2019everybody}. However, it was shown that this network’s structure performs better when the input and output are spatially aligned \cite{siarohin2018deformable}. This motivated the incorporation of the alignment into our scheme. 

We train our system in a supervised manner by actively pixelating high resolution video data. A reconstruction loss encourages adherence to the ground-truth, with an added perceptual loss offering further feature-level support. 
The addition of a discriminator assists in sharpening the outcome even further, thereby rounding off our setup. Using the aligned surrounding stack of frames also provides implicit temporal coherence for consecutive frames, and eliminates the need for an explicit temporal coherence loss.
We demonstrate the competence of our method in de-pixelating coarsely pixelated video clips, and compare to state-of-the-art single image facial SR and general video SR techniques. Figure \ref{fig:teaser} showcases four de-pixelization results of frames pixelated to a resolution of 16x16 and 8x8.

\noindent

\section{Related work}
\label{sec:related}

This work addresses video de-pixelization of human faces, and is therefore related to image and video super-resolution (SR) approaches, particularly those that focus on facial SR.

\textbf{\textit{Face Hallucination.}}
With the lion's share of SR solutions targeting general images, it is unsurprising that domain-specific SR has been mostly dedicated to human faces, 
our most prominent identifying feature. The main difference between the face hallucination problem and the generic image SR problem, is the fact that faces have unified structures which are highly familiar to humans.
As in the general domain, various methods, both classic \cite{baker2000hallucinating,liu2007face,tappen2012bayesian,yang2013structured} and CNN-based \cite{jin2015robust,kolouri2015transport,zhu2016deep,song2017learning}, have been proposed to address facial SR. 
{Crucial details are often missing from an LR image of a face, the recovery of which is an ill-posed problem, particularly in a single image setting. This difficulty is commonly handled with the use of additional priors, which are introduced to promote identity preservation. FSRNet}
\cite{chen2018fsrnet}
{synthesize a coarse HR image and improve it by estimating facial landmark heatmaps and parsing maps. Song}
\etal \cite{song2019joint}
{restore a coarse facial image using a CNN, and then utilize the exemplar-based detail enhancement algorithm via facial component matching.}

{Others apply multi-scale SR along with separate identity preservation using identity priors}
\cite{kazemi2019identity, grm2019face,zhu2016deep, hsu2019sigan}.
{Wavelet-SRNet}
\cite{huang2017wavelet}
{combines multi-level wavelet coefficient prediction to facilitate significant upscaling of highly pixelized facial images. Hu}
\etal \cite{hu2020face}
{estimate and extract 3D facial coefficients from an LR facial image, with which a sharp face rendered structure is created.
Other notable endeavors unify the tasks of face hallucination and face recognition in order to achieve realistic HR face image}
\cite{kong2019cross, bayramli2019fh}.
{Shi} 
\etal \cite{shi2019face} 
{introduce a deep RL-based optimization method to learn a series of ordered patch hallucination sequences. PULSE} 
\cite{menon2020pulse} 
{address the ill-posed problem of mapping an LR image to HR by searching in the latent space of a pre-trained StyleGAN. 
The main difference between these techniques and ours is that we utilize surrounding frames to recover fine details and promote identity preservation.}

\textbf{\textit{Video Super-Resolution.}}
Super-resolution of a video clip enjoys the plurality of frames providing further information useful for recovery of fine details, but must also exercise caution throughout the process, in order to maintain temporal coherence.
A selection of CNN-based methods have been proposed in recent years.
Liao \etal \cite{liao2015video} generate and combine sets of HR candidate patches from multiple frames. Others employ various forms of motion compensation \cite{kappeler2016video} and sub-pixel compensation \cite{tao2017detail,caballero2017real} to account for inter-frame differences. Liu \etal \cite{liu2017robust} combine an adaptive temporal aggregation mechanism with spatial alignment, Sajjadi \etal \cite{sajjadi2018frame} utilize a recurrent network (RNN) to minimize costly alignment operations by super-resolving each frame as a warp of its predecessor, and Wang \etal \cite{wang2019edvr} align frames at the feature-level in a coarse to fine manner, with a temporal and spatial attention fusion module emphasizing important features.   

{In facial video SR, Yoshida}
\etal \cite{Yoshida2012RobustFS}
{utilize classic techniques to combine image alignment and reconstruction. Deshmukh}
\etal \cite{deshmukh2019face}
{introduce an end-to-end facial video SR based on a CNN, operating on each frame in a sequence. In comparison, our method aggregates information from multiple frames for identity preservation and increased frame quality. Ataer-Cansizoglu}
\etal \cite{Cansizoglu2018SuperresolutionOV}
{super resolve each frame independently, and then generate weights to fuse frames together. This approach learns the fusion part but does not explicitly consider head movements and alignment of corresponding information.}

\section{Method}
\label{sec:method}

\begin{figure*}[h]
    \centering
    \captionsetup{labelfont=bf}
    \includegraphics[height=7cm]{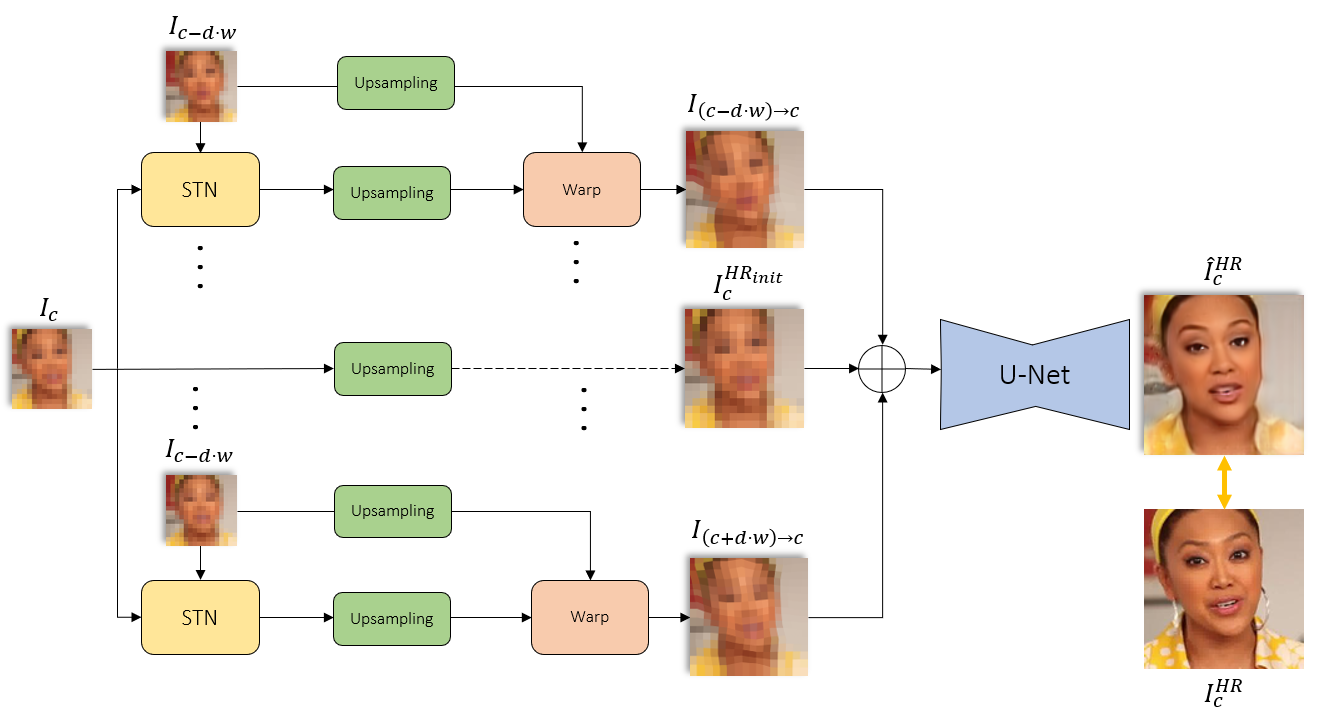}
    \caption{Method pipeline. The surrounding frames of the input $I_c$ are first aligned using a pretrained STN, and are then stacked in-order and jointly analyzed by the recovery U-Net component, which outputs the predicted HR frame $\hat{I}_c^{\text{HR}}$. During training, this frame is compared to the original high resolution frame $I_c^{\text{HR}}$.}
    \label{fig:archi}
\end{figure*}

Our system operates on pixelated video clips and processes each LR frame to obtain its HR version.
Each frame on its own provides very little information in its pixelated state, therefore frame recovery is carried out in stacks, where the frames surrounding a given image are analyzed alongside it.
Despite a high likelihood of temporal coherence along neighboring frames, misalignments are still present, and their extent only increases as we pull away from the current frame and widen our support window.

To handle these misalignments, we introduce a spatial transformation component (STN) \cite{jaderberg2015spatial} that is activated for any given frame, and precedes the de-pixelization component, as shown in Figure \ref{fig:archi}.
The STN receives as input two images, $I_A, I_B$, concatenated along their channel dimension, and outputs a free-form deformation grid $g_{B\rightarrow A}$ defining the warp from $I_B$ to $I_A$.

\begin{figure*}
\newcommand{\cmpfig}{1.4}
\newcommand{\cmpfigsp}{0.1} 
\setlength\tabcolsep{3pt} 
\renewcommand{\arraystretch}{0.5}
\captionsetup{labelfont=bf}
\centering
\begin{center}
\begin{tabular}{c c c | c c c}

A & B & B $\rightarrow$ A & A & B & B $\rightarrow$ A
\\ 
\\
\hline
\\
\includegraphics[height=\cmpfig cm]{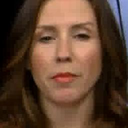} & 
\includegraphics[height=\cmpfig cm]{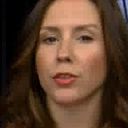}
& &
\includegraphics[height=\cmpfig cm]{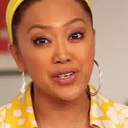} & 
\includegraphics[height=\cmpfig cm]{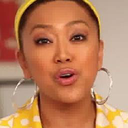} 
& \\
\includegraphics[height=\cmpfig cm]{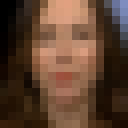} & 
\includegraphics[height=\cmpfig cm]{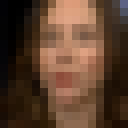} &
\includegraphics[height=\cmpfig cm]{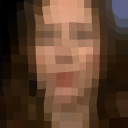} &
\includegraphics[height=\cmpfig cm]{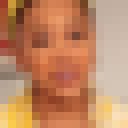} & 
\includegraphics[height=\cmpfig cm]{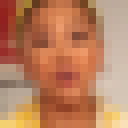} &
\includegraphics[height=\cmpfig cm]{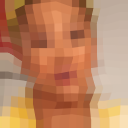} \\
\includegraphics[height=\cmpfig cm]{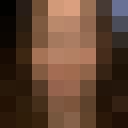} & 
\includegraphics[height=\cmpfig cm]{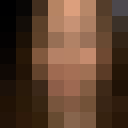} & 
\includegraphics[height=\cmpfig cm]{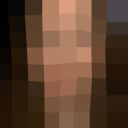} &
\includegraphics[height=\cmpfig cm]{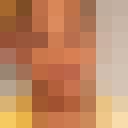} & 
\includegraphics[height=\cmpfig cm]{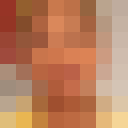} & 
\includegraphics[height=\cmpfig cm]{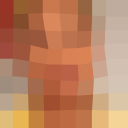} \\
\hline 
\end{tabular}
\end{center}
\vspace{-0.6cm}
\caption{Aligning pixelated frames using our STN. In each example we present two frames from the same clip (A and B) and their pixelized versions (16x16 and 8x8). The third column in each example displays the warped version of B to fit A.}
\label{fig:align}
\end{figure*}

We adopt a U-Net \cite{ronneberger2015u} architecture for our STN, where the LR input frames and the requested free-form grid are of a similar resolution, 
but not necessarily identical. In our experiments, we set the warp grid resolution to 8x8, and the input frame resolution to either 8x8 or 16x16 (see Figure \ref{fig:align}).
We opted for a relatively large grid size that is expressive enough to support a wide range of possible deformations among frames. 

Having attempted to train the STN and the de-pixelization components end-to-end, we experienced difficulties with balancing between the two, with the STN exhibiting substantially faster convergence that ultimately led to over-fitting. Hence, since the alignment and SR tasks are conceptually separate, there is no requirement to train them in a unified pipeline, and the STN is trained prior to the de-pixelization training, on pairs of frames that are drawn from the same video clip.
Its loss contains a reconstruction element ($\mathcal{L}_R^{\text{STN}}$) that penalizes the distance between $I_A$ and the warped $I_B$, and a regularization element ($\mathcal{L}_I^{\text{STN}}$) that prevents the grid from straying too far from the identity warp grid:

\begin{equation}
    \mathcal{L}_R^{\text{STN}}=\text{L1}(I_A,g^*_{B\rightarrow A}(I_B))
\label{eq:stn_recon}
\end{equation}

\begin{equation}
    \mathcal{L}_I^{\text{STN}}=\text{L1}(g_{\text{id}},g_{B\rightarrow A})
\label{eq:stn_id}
\end{equation}

In Equation \ref{eq:stn_recon}, $g^*_{B\rightarrow A}$ is the warp grid computed by the STN, upsampled (using bilinear interpolation) to the corresponding image resolution (when grid resolution equals image resolution, we get $g^*=g$), thus $g^*_{B\rightarrow A}(I_B)$ denotes the result of applying the warp defined by $g_{B\rightarrow A}$ on $I_B$. In Equation \ref{eq:stn_id}, $g_{\text{id}}$ denotes the identity warp grid. See Figure \ref{fig:align} for STN alignment examples on pixelated frames.

The stack per given frame $I_c$ is created by using $2w$ images sampled from a symmetric window around frame $c$. We use a spacing parameter $d$ and extract all frames $I_j$ where $j = \{c+dj\}_{j=-w}^{w}$. We align all of these frames to $I_c$ using our trained STN.
Each such activation yields a warp grid $g_{j\rightarrow c}$ that is upsampled to our desired HR resolution, and applied onto its corresponding supporting frame $I_j$, that is also upsampled (using bicubic interpolation) to the same resolution. The stack of upsampled and warped frames are then concatenated along their color dimension, with an upsampled $I_c$ situated in the middle.

The stack of frames is inserted into our de-pixelization network, which is a simple U-Net based encoder-decoder. The structure of this network and its usage of skip connections, utilizes and benefits from the strong alignment that characterizes the data. 
The output of the network is a single image $\hat{I}_c^{\text{HR}}$, which is the predicted HR version of the middle frame $I_c$.
This network employs a standard reconstruction loss ($\mathcal{L}_R$), comparing $\hat{I}_c^{\text{HR}}$ to its groundtruth HR frame $I_c^{\text{HR}}$, and a perceptual similarity loss ($\mathcal{L}_P$) that compares the deep feature maps of the two:

\begin{equation}
    \mathcal{L}_R=\text{L1}(\hat{I}_c^{\text{HR}},I_c^{\text{HR}})
\label{eq:recon_loss}
\end{equation}

\begin{equation}
    \mathcal{L}_P=\text{L1}(\text{FM}(\hat{I}_c^{\text{HR}}),\text{FM}(I_c^{\text{HR}}))
\label{eq:perc_loss}
\end{equation}

In Equation \ref{eq:perc_loss}, $\text{FM}$ denotes a feature map. In our experiments, we use VGG-19 \cite{simonyan2014very} (layers 1, 6, 11, 20, 29), and VGG-Face \cite{parkhi2015deep} (layers 1, 6, 11, 18, 25). See Figure \ref{fig:archi} for an illustration of our pipeline.

As a final boost, we add a discriminator which helps to produce sharper images that appear more real and natural. We use a PatchGAN discriminator \cite{li2016precomputed} and standard adversarial losses, and feed it pairs of images made up of a pixelated frame and its HR version, with groundtruth frames as \textit{real} and output frames as \textit{fake}.

\section{Evaluation}
\label{sec:results}
In this section we perform a quantitative and qualitative evaluation of our proposed approach. We conduct an ablation study to explore the contribution of each of our design choices, and compare to single image facial SR and general video SR methods. 

\begin{figure*}
\captionsetup{labelfont=bf}
\newcommand{\cmpfig}{1.7}
\newcommand{\cmpfigsp}{0.05} 
\setlength\tabcolsep{1pt} 
\centering
\begin{center}

\begin{tabular}{ 
>{\centering\arraybackslash}m{0.7in} >{\centering\arraybackslash}m{0.7in}
>{\centering\arraybackslash}m{0.7in} >{\centering\arraybackslash}m{0.7in}
>{\centering\arraybackslash}m{0.7in}  >{\centering\arraybackslash}m{0.7in}}

(1) & (2) & (3) & (4) & (5) & (6)\\

\includegraphics[height=\cmpfig cm]{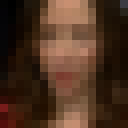} &
\includegraphics[height=\cmpfig cm]{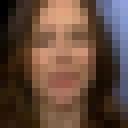} & \includegraphics[height=\cmpfig cm]{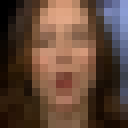} & \includegraphics[height=\cmpfig cm]{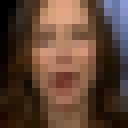} & \includegraphics[height=\cmpfig cm]{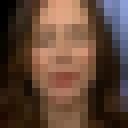} &
\includegraphics[height=\cmpfig cm]{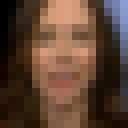} \\

\includegraphics[height=\cmpfig cm]{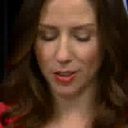} &
\includegraphics[height=\cmpfig cm]{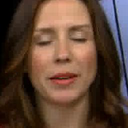} & \includegraphics[height=\cmpfig cm]{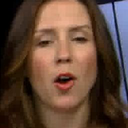} & \includegraphics[height=\cmpfig cm]{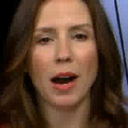} & \includegraphics[height=\cmpfig cm]{figures/results/images/308_68gt_85.png} &
\includegraphics[height=\cmpfig cm]{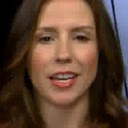} \\

\includegraphics[height=\cmpfig cm]{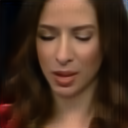} &
\includegraphics[height=\cmpfig cm]{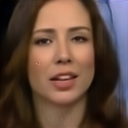}  &
\includegraphics[height=\cmpfig cm]{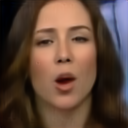}  &
\includegraphics[height=\cmpfig cm]{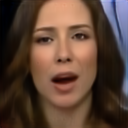}  &
\includegraphics[height=\cmpfig cm]{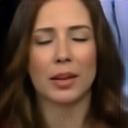} &
\includegraphics[height=\cmpfig cm]{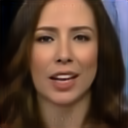} \\




\includegraphics[height=\cmpfig cm]{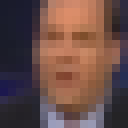} &
\includegraphics[height=\cmpfig cm]{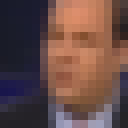}  &
\includegraphics[height=\cmpfig cm]{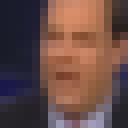}  &
\includegraphics[height=\cmpfig cm]{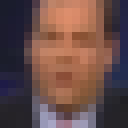}  &
\includegraphics[height=\cmpfig cm]{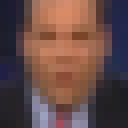} &
\includegraphics[height=\cmpfig cm]{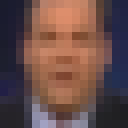} \\

\includegraphics[height=\cmpfig cm]{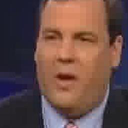} &
\includegraphics[height=\cmpfig cm]{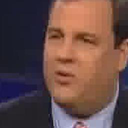}  &
\includegraphics[height=\cmpfig cm]{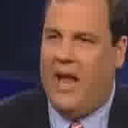}  &
\includegraphics[height=\cmpfig cm]{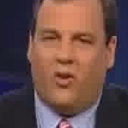}  &
\includegraphics[height=\cmpfig cm]{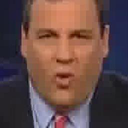} &
\includegraphics[height=\cmpfig cm]{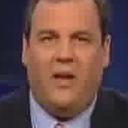} \\

\includegraphics[height=\cmpfig cm]{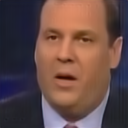} &
\includegraphics[height=\cmpfig cm]{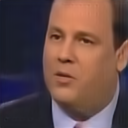}  &
\includegraphics[height=\cmpfig cm]{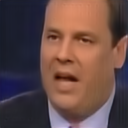}  &
\includegraphics[height=\cmpfig cm]{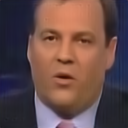} &
\includegraphics[height=\cmpfig cm]{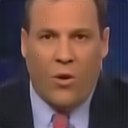} &
\includegraphics[height=\cmpfig cm]{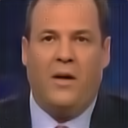} \\

\end{tabular}
\caption{A selection of our results. We present a subset of six frames from three clips, where the first row in each displays the 16x16 pixelated input, the second row contains the groundtruth frames, and the third - the 128x128 output of our pipeline.}
\label{fig:res}
\end{center}
\end{figure*}

Figure \ref{fig:res} showcases a selection of our results
demonstrating that our method generates high-quality, realistic faces with identifying features that remain faithful to those of the underlying subjects. Full video clips are available in our supplementary material.

Throughout this section, we utilize the widely recognized PSNR and SSIM measures to quantitatively evaluate our performance vs. ablated versions and compared techniques.
PSNR computes the mean squared reconstruction error after denoising, where a higher value indicates a better result. By measuring a least squares error (on low frequency content), it creates a bias toward overly smoothed images, which often appear non-realistic. On the other hand, SSIM takes into account edge similarity (high frequency content) between output and groundtruth.
Additionally, to measure identity similarity, we compute the cosine distance between deep feature maps extracted from VGG-Face \cite{parkhi2015deep}.

\subsection{Dataset and training}
\label{sub:dataset}
We perform our evaluation on the FaceForensics++ dataset \cite{rossler2019faceforensics++}, containing 977 YouTube video clips of people talking to the camera.
To prepare the data for our needs, we extract the frames of each clip, and crop out the face using Bulat et al. \cite{bulat2017far}. The head crop is saved at a resolution of 256x256. We generate a coarse pixelated version of each frame by downsampling to a desired target resolution (8x8 or 16x16 in our experiments), and then upsampling back to a high resolution (128x128 in our experiments) using nearest-neighbor interpolation. As such, for each video, we possess its constituent frames as pairs of pixelated and groundtruth HR images. 

All our experiments were conducted on a separate test set comprised of $10\%$ of the FaceForensics++ dataset.

We train our pipline with a batch size of 16, and set $w=2, d=5$, thus the input to our de-pixelization component is of size 16x15x128x128, since the stack of images per input frame amounts to 15 channels. Please refer to our supplementary material for further information regarding our architecture and implementation details.

\subsection{Ablation}
\label{sub:abl}
Our full pipeline is composed of an STN component, a de-pixelization component, and a discriminator. In this ablation study, we compare the quantitative and qualitative performance of our full solution, to two partial versions, the first excludes the STN, and the second the adversarial setting (no discriminator).

\paragraph{\textbf{STN.}}
In this version, de-pixelization is trained with unaligned pixelated frames, \textit{i.e.,}, without using the pretrained STN to predict alignment warp fields. Example results are shown in {Figure ~\ref{fig:abl} (Right)} and in the supplementary video.
Note the difference in overall quality between the full solution (fourth column) and the ablated version (third column). The full version produces images with a lesser degree of blurriness, while the ablated features various disturbing artifacts.
 
Our full pipeline better preserves facial expressions, as demonstrated in {Figure~\ref{fig:abl} (Right)}. In the first row, notice the differences in mouth pose, and in the third row, the blinking state of the right eye.

Personal identifying features are also better recovered by the full solution, where, for instance, the slight curvature of the cheekbone of the woman in the second row is smoothed out in the ablated version, and the bridge of the nose of the woman in the fourth row is narrower and more similar to the groundtruth, in the full version.

\begin{figure*}[h]
\captionsetup{labelfont=bf}
\newcommand{\cmpfig}{1.7}
\newcommand{\cmpfigsp}{0.1} 
\setlength\tabcolsep{1pt} 
\newcommand{\plfig}{7}

\begin{subfigure}{.5\textwidth}
    \centering
    \begin{tabular}{ >{\centering\arraybackslash}m{0.35in} >{\centering\arraybackslash}m{0.7in}  >{\centering\arraybackslash}m{0.7in} >{\centering\arraybackslash}m{0.7in} >{\centering\arraybackslash}m{0.7in}|}
    & input & GT & w/o disc. & ours \\
    (1) &
    \includegraphics[height=\cmpfig cm]{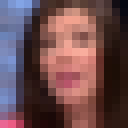} &
    \includegraphics[height=\cmpfig cm]{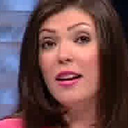} &
    \includegraphics[height=\cmpfig cm]{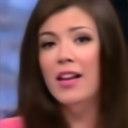} &
    \includegraphics[height=\cmpfig cm]{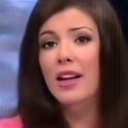} \\
    
    (2) &
    \includegraphics[height=\cmpfig cm]{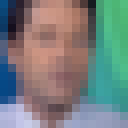} &
    \includegraphics[height=\cmpfig cm]{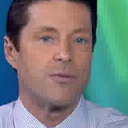} &
    \includegraphics[height=\cmpfig cm]{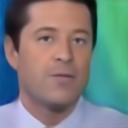} &
    \includegraphics[height=\cmpfig cm]{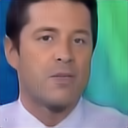} \\
    
    (3) &
    \includegraphics[height=\cmpfig cm]{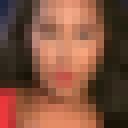} &
    \includegraphics[height=\cmpfig cm]{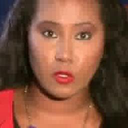} &
    \includegraphics[height=\cmpfig cm]{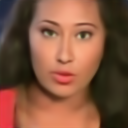} &
    \includegraphics[height=\cmpfig cm]{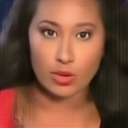} \\
    
    (4) &
    \includegraphics[height=\cmpfig cm]{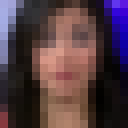} &
    \includegraphics[height=\cmpfig cm]{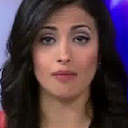} &
    \includegraphics[height=\cmpfig cm]{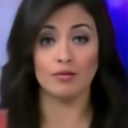} &
    \includegraphics[height=\cmpfig cm]{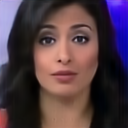} \\
    \end{tabular}
    \label{fig:abl_GAN}
\end{subfigure}
\begin{subfigure}{.5\textwidth}
    \centering
    \input{figures/ablation/abl_STN}
\end{subfigure}
\caption{Qualitative ablation study. Left: GAN - The ablated version (third column) produces images that are overly smooth, with inattention to fine details. Note the deformation of the left eye of both of the women in the third and fourth rows. Right: STN - The ablated version (third column) produces images with various artifacts and that are generally more blurry, and is less faithful to the expressions and identifying personal features of the subjects.}
\label{fig:abl}
\end{figure*}

\paragraph{\textbf{Adversarial.}} 
In this version, our pipeline is trained with the exclusion of the discriminator (no adversarial losses). See Figure  {Figure~\ref{fig:abl} (Left)} for several examples demonstrating that the ablated version produces overly smooth images lacking fine details. In some cases, facial features are unnaturally distorted, for instance, note the left eye of both of the women in the third and fourth rows. These results are in-line with the claim-to-fame of the GAN paradigm, promoting synthesized image realism and sharpness.

Table \ref{tab:tbl_abl} summarizes the quantitative results of our ablation experiment, where our full solution achieves the best scores in all measures.
Identity similarity clearly increases when we incorporate the STN component, but it is less affected by the inclusion or exclusion of the adversary, since the adversarial setting targets general domain realism.
PSNR is similarly minimally affected by the existence of the adversary, due to its low sensitivity to changes in the high frequency domain. Conversely, SSIM emphasizes edge similarity, and is therefore more dramatically affected by the existence of the adversary.

\begin{table}
\captionsetup{labelfont=bf}
\centering
    \begin{tabular}{|| c || c c c ||} 
     \hline
     version & PSNR & SSIM & ID \\ [1ex] 
     \hline\hline
     
     w/o STN &  25.0829 &  0.7643 & 0.8768 \\ 
     \hline
     w/o disc. &  26.2443 &  0.7665 &   0.8846 \\ 
     \hline
     full &  \textbf{26.3309} &  \textbf{0.7866} &  \textbf{0.8867} \\  
     \hline
     
    \end{tabular}

    \centering
    \caption{\label{tab:tbl_abl}Quantitative ablation results. We compare our full solution (bottom row) against two ablated versions - one without STN (first row) and the other without an adversary (second row). The last column (ID) is the identity similarity computed by cosine distance in VGG-Face feature space.}
\end{table}

\subsection{Comparisons}
\label{sub:comp}
\begin{figure}[H]
    \captionsetup{labelfont=bf, oneside ,margin={0.5cm,-2cm}}
    \newcommand{\cmpfig}{1.7}
    \newcommand{\cmpfigsp}{0.2} 
    \setlength\tabcolsep{1pt} 
    \centering
    \begin{center}
    \begin{tabular}{ >{\centering\arraybackslash}m{0.3in} | >{\centering\arraybackslash}m{0.7in} >{\centering\arraybackslash}m{0.7in}| >{\centering\arraybackslash}m{0.7in} >{\centering\arraybackslash}m{0.7in} || >{\centering\arraybackslash}m{0.7in}|}
    & input & GT & W-SRNet & EDVR & ours \\
    (1) &
    \includegraphics[height=\cmpfig cm]{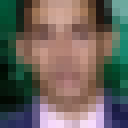} &
    \includegraphics[height=\cmpfig cm]{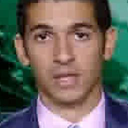} &
    \includegraphics[height=\cmpfig cm]{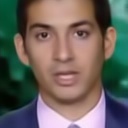} &
    \includegraphics[height=\cmpfig cm]{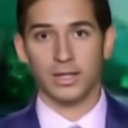} &
    \includegraphics[height=\cmpfig cm]{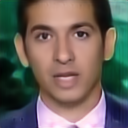} \\
    
    (2) &
    \includegraphics[height=\cmpfig cm]{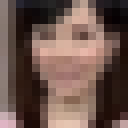} &
    \includegraphics[height=\cmpfig cm]{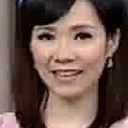} &
    \includegraphics[height=\cmpfig cm]{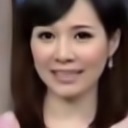} &
    \includegraphics[height=\cmpfig cm]{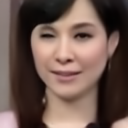} &
    \includegraphics[height=\cmpfig cm]{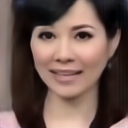} \\
    
    (3) &
    \includegraphics[height=\cmpfig cm]{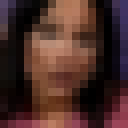} &
    \includegraphics[height=\cmpfig cm]{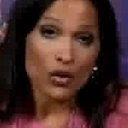} &
    \includegraphics[height=\cmpfig cm]{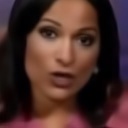} &
    \includegraphics[height=\cmpfig cm]{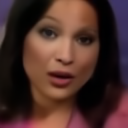} &
    \includegraphics[height=\cmpfig cm]{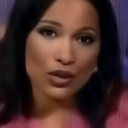} \\

    (4) &
    \includegraphics[height=\cmpfig cm]{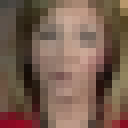} &
    \includegraphics[height=\cmpfig cm]{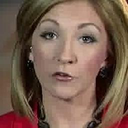} &
    \includegraphics[height=\cmpfig cm]{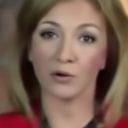} &
    \includegraphics[height=\cmpfig cm]{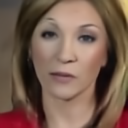} &
    \includegraphics[height=\cmpfig cm]{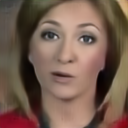} \\
    
    (5) &
    \includegraphics[height=\cmpfig cm]{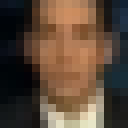} &
    \includegraphics[height=\cmpfig cm]{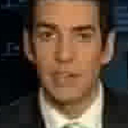} &
    \includegraphics[height=\cmpfig cm]{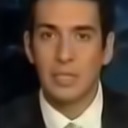} &
    \includegraphics[height=\cmpfig cm]{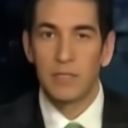} &
    \includegraphics[height=\cmpfig cm]{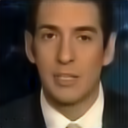} \\
    
    (6) &
    \includegraphics[height=\cmpfig cm]{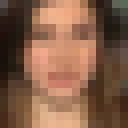} &
    \includegraphics[height=\cmpfig cm]{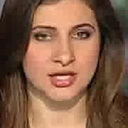} &
    \includegraphics[height=\cmpfig cm]{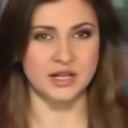} &
    \includegraphics[height=\cmpfig cm]{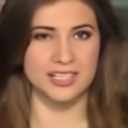} &
    \includegraphics[height=\cmpfig cm]{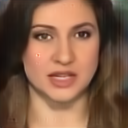} \\
    \end{tabular}
    
    \caption{Qualitative comparisons. We compare our solution (last column) against Wavelet SRNet - a single image facial SR method (third column), and EDVR - a general video SR method (fourth column).}
    \label{fig:comp}
    \end{center}
\end{figure}

We compare our solution to two state-of-the-art techniques, and summarize the results in Table \ref{tab:tbl_comp} and Figure \ref{fig:comp}. Both techniques were trained on the same training set as ours, using the official implementations and settings.
\paragraph{\textbf{Wavelet-SRNet}} \cite{huang2017wavelet} is a single image facial SR approach that learns to reconstruct wavelet coefficients from an LR input, that are then used to recover the HR version of the image.
This approach leverages wavelets to encode texture.
We trained Wavelet-SRNet to super-resolve 16x16 pixelized inputs, to a resolution of 128x128.
Examining their results in Figure \ref{fig:comp} (third column), we note the high plausibility of the output and faithfulness to the underlying identity. Although their method uses explicit facial priors, we observe that our method (fourth column) generates overall sharper results with an increased level of realism, and better reconstructs facial expressions (for example, third and fourth rows).
\paragraph{\textbf{EDVR}} \cite{wang2019edvr} is a general video SR technique combining spatial and temporal attention with feature-level alignment. EDVR is targeted toward x4 upscaling, thus was trained on 32x32 resolution frames that were upsampled from a resolution of 16x16 using bicubic interpolation, and was asked to output 128x128 frames. The results appear in Figure \ref{fig:comp} (fourth column), and demonstrate the least adherence to the groundtruth in terms of identity recovery (rows 1,2,3,6), as well as facial expression reconstruction (rows 2,4,5,6). 

The quantitative results appear in Table \ref{tab:tbl_abl}. 
Our proposed method obtains the highest scores in all measures. Our network is explicitly trained to minimize the perceptual distance between the output and the groundtruth, which contributes to its increased performance. In comparison, despite having trained EDVR on our data, which is restricted to human faces, it is designed for general video SR, and is therefore less suitable for the specific task of facial de-pixelization. Wavelet-SRNet is indeed designed specifically for human face SR, but operates on single images,
thus the plurality of frames within a video is not exploited.

Lastly, we report that training our full pipeline (with $w=2$) took a total of 50 hours on a single NVIDIA GeForce GTX 1080 GPU, whereas Wavelet-SRNet trained for 12.5 days (single), and EDVR trained for 7 days on two GPUs.
\begin{table}
\captionsetup{labelfont=bf}
\centering
 \begin{tabular}{||c || c c c||} 
     \hline
     method & PSNR & SSIM & ID \\ [0.5ex] 
     \hline\hline
     W-SRNet & 26.169 & 0.7813 & 0.7051  \\ 
     \hline
     EDVR & 24.604 & 0.7814 & 0.6941 \\
     \hline
     ours & \textbf{26.330} & \textbf{0.7866} & \textbf{0.8867} \\
     \hline
\end{tabular}
\centering
\caption{\label{tab:tbl_comp}Quantitative comparison results. We compare our solution against Wavelet SRNet - a single image facial SR method, and EDVR - a general video SR method.}
\end{table}

\subsection{Stress test}
\begin{figure}[H]
\captionsetup{labelfont=bf, oneside ,margin={2.5cm,0.5cm}}
\newcommand{\cmpfig}{1.7}
\newcommand{\cmpfigsp}{0.2} 
\setlength\tabcolsep{1pt} 
\centering

\hspace{2.5cm}
\begin{tabular}{|| >{\centering\arraybackslash}m{0.7in} >{\centering\arraybackslash}m{0.7in} |
>{\centering\arraybackslash}m{0.7in}}

input & GT & output \\
\includegraphics[height=\cmpfig cm]{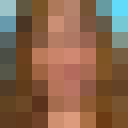}  &
\includegraphics[height=\cmpfig cm]{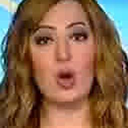} &
\includegraphics[height=\cmpfig cm]{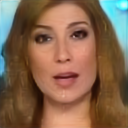} \\
\includegraphics[height=\cmpfig cm]{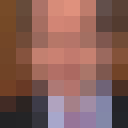} &
\includegraphics[height=\cmpfig cm]{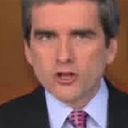} &
\includegraphics[height=\cmpfig cm]{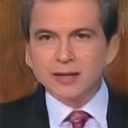} \\

\includegraphics[height=\cmpfig cm]{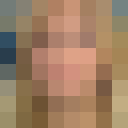}  &
\includegraphics[height=\cmpfig cm]{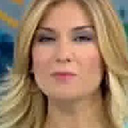} &
\includegraphics[height=\cmpfig cm]{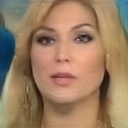} \\
\includegraphics[height=\cmpfig cm]{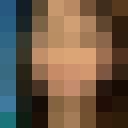} &
\includegraphics[height=\cmpfig cm]{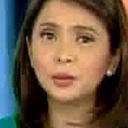} &
\includegraphics[height=\cmpfig cm]{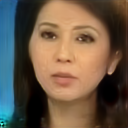} \\
\includegraphics[height=\cmpfig cm]{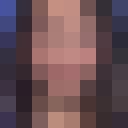} &
\includegraphics[height=\cmpfig cm]{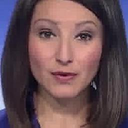} &
\includegraphics[height=\cmpfig cm]{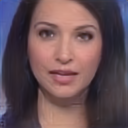} \\

\includegraphics[height=\cmpfig cm]{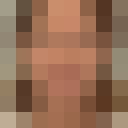} &
\includegraphics[height=\cmpfig cm]{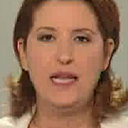} &
\includegraphics[height=\cmpfig cm]{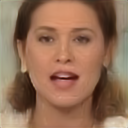} \\
\end{tabular}
\caption{Stress test examples. Input frames are pixelated to a resolution of 8x8, and de-pixelized by our system to a resolution of 128x128.}
\label{fig:pix8}
\end{figure}

In this experiment, we train our system as stated above, but use an 8x8 pixelization instead of 16x16.
Figure \ref{fig:pix8} demonstrates that even under such coarse pixelization, our method succeeds in hallucinating plausible faces with facial expressions that are similar to the groundtruth. Naturally, in comparison to the 16x16 version, identity preservation is compromised, but the generated images depict subjects that correlate well with the highly pixelated inputs.


\subsection{Support window size}
\begin{figure}
    \captionsetup{labelfont=bf}
    \centering
    \includegraphics[height=5.0cm]{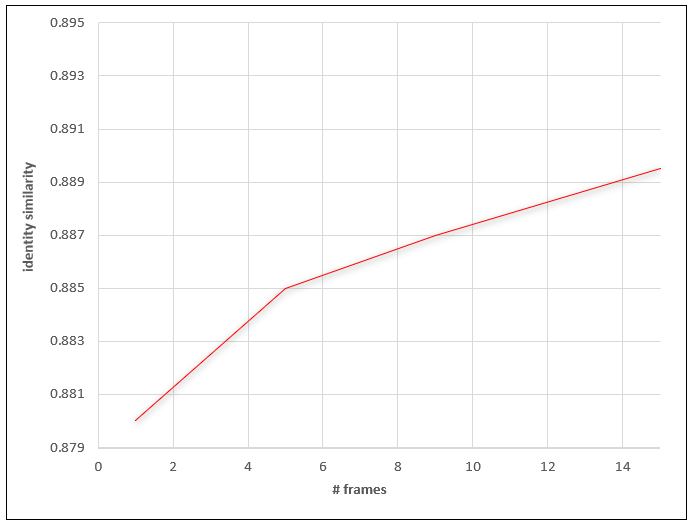}
    \caption{Support window size impact. Identity similarity score as a function of frame stack size.}
    \label{graph:id_frames}
\end{figure}

\begin{figure}[H]
\newcommand{\cmpfig}{1.4}
\newcommand{\cmpfigsp}{1.1} 
\captionsetup{labelfont=bf}

\hspace{-5mm}
\begin{tabular}{>{\centering\arraybackslash}m{\cmpfigsp cm} >{\centering\arraybackslash}m{1.2 cm} |  >{\centering\arraybackslash}m{\cmpfigsp cm} >{\centering\arraybackslash}m{\cmpfigsp cm}  >{\centering\arraybackslash}m{\cmpfigsp cm} >{\centering\arraybackslash}m{\cmpfigsp cm}}
    input & GT & 1f & 5f & 9f & 15f\\
    \includegraphics[height=\cmpfig cm]{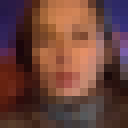} &
    \includegraphics[height=\cmpfig cm]{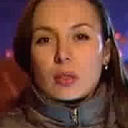} &
    \includegraphics[height=\cmpfig cm]{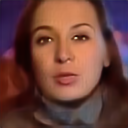} & 
    \includegraphics[height=\cmpfig cm]{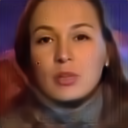} &
    \includegraphics[height=\cmpfig cm]{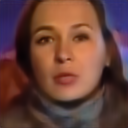} &
    \includegraphics[height=\cmpfig cm]{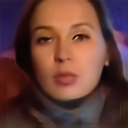} \\
    \includegraphics[height=\cmpfig cm]{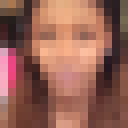} &
    \includegraphics[height=\cmpfig cm]{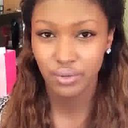} &
    \includegraphics[height=\cmpfig cm]{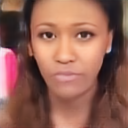} & 
    \includegraphics[height=\cmpfig cm]{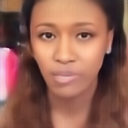} &
    \includegraphics[height=\cmpfig cm]{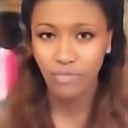} &
    \includegraphics[height=\cmpfig cm]{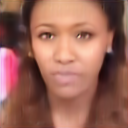} \\
\end{tabular}
\caption{We progressively increase support window size and explore its effect on generated frame quality and identity recovery.}
\label{fig:temp}
\end{figure}

We train our pipeline with various support window sizes to explore the impact of providing narrow to wide support from surrounding frames. Figure \ref{fig:temp} presents visual comparisons of recovered frames under our selected set of sizes: $F=1,5,9,15$, where $F$ is the size of the input stack (recall that $F=2w+1$),
and Figure \ref{graph:id_frames} plots the identity similarity score as a function of $F$. We note the dramatic improvement achieved by increasing the stack size from 1 (single frame operation) to 5, indicating that aggregating information from multiple surrounding frames assists in recovering identifying features (see examples in Figure \ref{fig:temp}, for instance, the eyes and lips of the woman in the second row). We also note, that while the score keeps rising as stack size is further increased, its impact starts to weaken. As we observe in Figure \ref{fig:temp}, generated frames become increasingly blurry as we widen the support window, suggesting that there is a tradeoff between identity preservation and overall image quality. The latter is clearly compromised when frames within an input stack become too numerous to provide consistent cues for reconstruction.

\section{Conclusions}
\label{sec:discussion}
De-pixelizing an image or a video is an ill-posed problem, since a substantial portion of the information is lost in the pixelization process, particularly higher frequency details that are crucial for face identification. As we have shown, surprisingly, it is still possible to recover a plausible and relatively similar face to the one encoded. We have shown that de-pixelizing a frame can greatly benefit from the information encapsulated in its surrounding neighbors, and furthermore, when they are properly aligned.

\begin{figure}[H]
\captionsetup{labelfont=bf}
\newcommand{\cmpfig}{1.7}
\newcommand{\cmpfigsp}{0.2} 
\setlength\tabcolsep{1pt} 
\centering
\begin{center}
\begin{tabular}{ 
>{\centering\arraybackslash}m{0.7in} >{\centering\arraybackslash}m{0.7in}
>{\centering\arraybackslash}m{0.7in} 
}
input & GT & output \\
\includegraphics[height=\cmpfig cm]{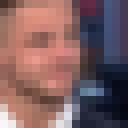} &
\includegraphics[height=\cmpfig cm]{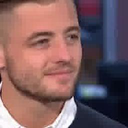}  &
\includegraphics[height=\cmpfig cm]{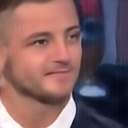} \\

\includegraphics[height=\cmpfig cm]{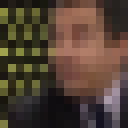} &
\includegraphics[height=\cmpfig cm]{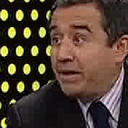}  &
\includegraphics[height=\cmpfig cm]{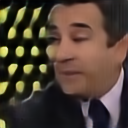} \\

\includegraphics[height=\cmpfig cm]{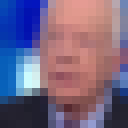} &
\includegraphics[height=\cmpfig cm]{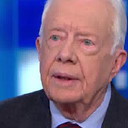}  &
\includegraphics[height=\cmpfig cm]{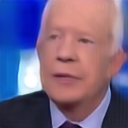} \\

\end{tabular}
\vspace{-3mm}
\caption{Failure cases. 
Our system struggles with pixelated faces under extreme poses (first and second rows), and may suffer when there are misleading background cues (second row). In addition, fine textural details, such as wrinkles, are harder to reconstruct (bottom row).}
\label{fig:fail}
\end{center}
\end{figure}

Our technique is not faultless. Figure~\ref{fig:fail} demonstrates a few failure cases, 
where extreme or unique head poses, confusing background patterns, and fine textural details such as wrinkles, 
induce synthesis of 
blurry output with distorted facial features.
As we have demonstrated, our technique can de-pixelize the frames of a video clip and generate a smooth and stable stream without directly enforcing temporal coherence. Since the input video is temporally coherent, the overlaps of the sliding window along the computation, naturally award a similar coherence to the output video. 
Our motivation, however, is to merely show that a pixelated video is not successful in concealing a subject's identity. In the future, we would like to investigate and develop means to apply a more resilient pixelization method. 
Possible directions include injection of noise or randomness into the process, and usage of a structured adversarial bias to prevent successful de-pixelization.

\clearpage
{\small
\bibliographystyle{ieee_fullname}
\bibliography{egbib}
}

\clearpage
\includepdf[pages=1]{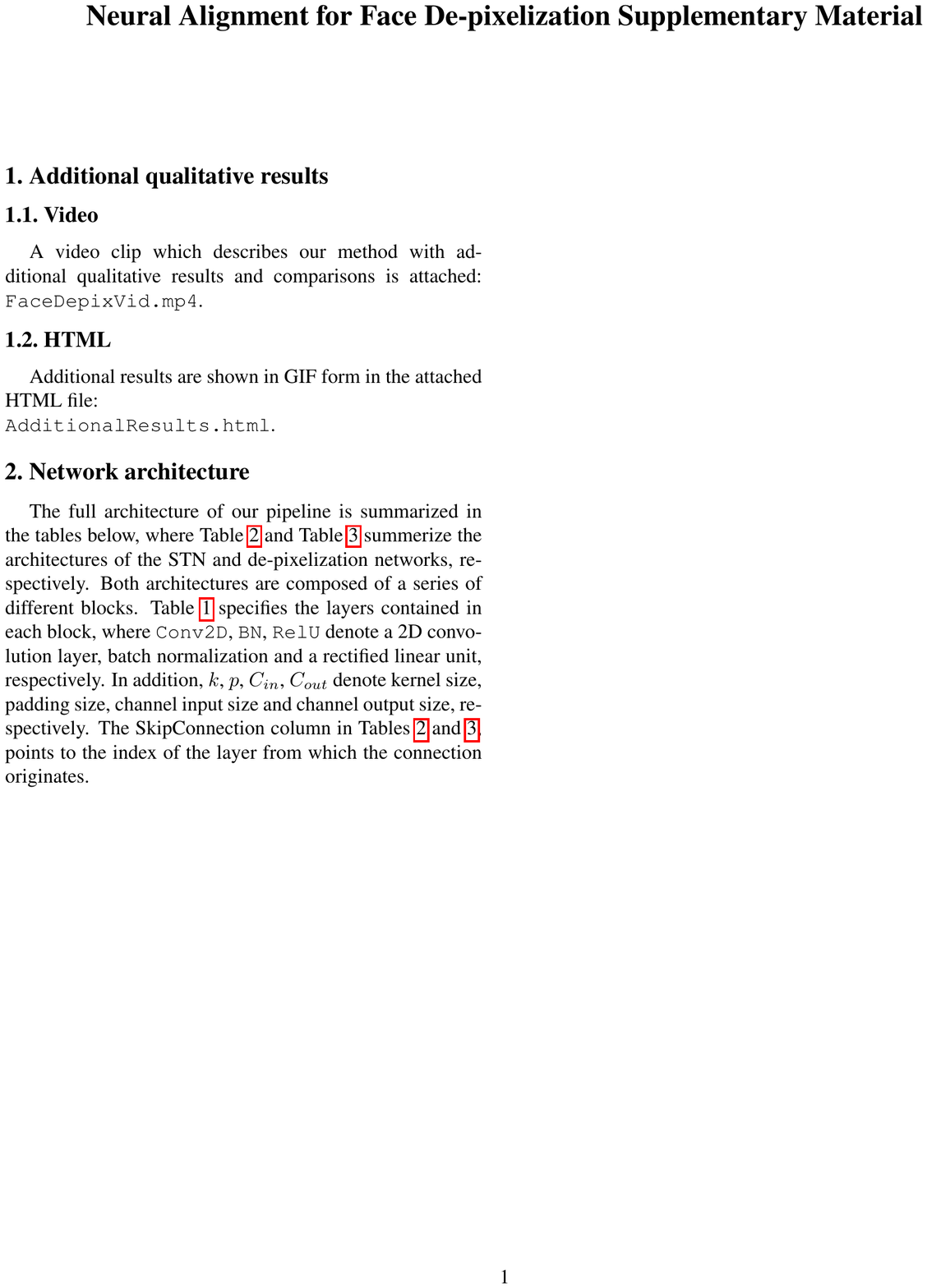}
\includepdf[pages=2]{07_supp.pdf}

\end{document}